\pdfoutput=1
\documentclass[11pt]{article}

\usepackage{acl}
\usepackage{times}
\usepackage{latexsym}
\usepackage[T1]{fontenc}
\usepackage[utf8]{inputenc}
\usepackage{microtype}
\usepackage{inconsolata}

\usepackage{xspace}
\usepackage{graphicx}
\usepackage{amsmath}
\usepackage{tabularx}
\usepackage{enumitem}
\usepackage{mdframed}
\usepackage{xcolor,colortbl}
\usepackage{url}
\usepackage{booktabs}

\newcommand\ours{SCM\xspace}

\newcommand{\scmturbo}{\ours\textsubscript{turbo}\xspace}
\newcommand{\scmdavinci}{\ours\textsubscript{davinci003}\xspace}

\newcommand{\figref}[1]{Figure~\ref{fig:#1}}
\newcommand{\tabref}[1]{Table~\ref{tab:#1}}
\newmdenv[
  backgroundcolor=red!05,
  linecolor=quoteborder,
  skipabove=1em,
  skipbelow=0em,
  leftline=true,
  topline=false,
  bottomline=false,
  rightline=false,
  linecolor=red!66,
  linewidth=4pt
]{githubquote}

\title{Enhancing Large Language Model with Self-Controlled Memory Framework}

\author{Bing Wang\textsuperscript{1}, Xinnian Liang\textsuperscript{1}, Jian Yang\textsuperscript{1}, Hui Huang\textsuperscript{2} \\
{\bf Shuangzhi Wu\textsuperscript{3}, Peihao Wu\textsuperscript{3}, Lu Lu\textsuperscript{3}, Zejun Ma\textsuperscript{3}  Zhoujun Li\textsuperscript{1}}\\ 
\textsuperscript{1}State Key Lab of Software Development Environment, Beihang University, Beijing, China \\ 
\textsuperscript{2}Harbin Institute of Technology, Harbin, China \\
\textsuperscript{3}ByteDance AI Lab, Beijing, China\\ 
\texttt{\{bingwang,xnliang,jiaya,lizj\}@buaa.edu.cn} \\
\texttt{ huanghui@stu.hit.edu.cn } \\
\texttt{\{wufurui,mazejun\}@bytedance.com}\\ }

\begin{document}
\maketitle

\renewcommand{\thefootnote}{\fnsymbol{footnote}} 
\renewcommand{\thefootnote}{\arabic{footnote}} % 切换回阿拉伯字母

\begin{abstract}
Large Language Models (LLMs) are constrained by their inability to process lengthy inputs, resulting in the loss of critical historical information. To address this limitation, in this paper, we propose the Self-Controlled Memory (SCM) framework to enhance the ability of LLMs to maintain long-term memory and recall relevant information. Our SCM framework comprises three key components: \textit{an LLM-based agent} serving as the backbone of the framework, \textit{a memory stream} storing agent memories, and \textit{a memory controller} updating memories and determining when and how to utilize memories from memory stream. Additionally, the proposed SCM is able to process ultra-long texts without any modification or fine-tuning, which can integrate with any instruction following LLMs in a plug-and-play paradigm. Furthermore, we annotate a dataset to evaluate the effectiveness of SCM for handling lengthy inputs. The annotated dataset covers three tasks: long-term dialogues, book summarization, and meeting summarization. Experimental results demonstrate that our method achieves better retrieval recall and generates more informative responses compared to competitive baselines in long-term dialogues. \footnote{ \url{https://github.com/wbbeyourself/SCM4LLMs}}
\end{abstract}

\section{Introduction}

Recently, Large Language Models (LLMs) have attracted significant attention due to their remarkable performance in various tasks~\cite{gpt3-openai,zeng2023glm-130b,ouyang2022training,thoppilan2022lamda}. 
% The strong foundational capability of LLMs, achieved through large-scale pre-training on massive text corpora (e.g., In-Context Learning~\cite{gpt3-openai}, Chain-of-Thoughts~\cite{wei2022chain,wei2022emergent}, among others), is a contributing factor to their success.
Instruction-tuning~\cite{2020t5,wei2022finetuned,chung2022scaling} helps LLMs comprehend natural language task descriptions, while reinforcement learning with human feedback~\cite{schulman2017proximal,summ-rlhf,bai2022training} aligns generated text with human preferences. 
% The combined capabilities of LLMs have effectively shattered the boundaries between natural language processing tasks, leading to limitless possibilities in the application and research directions of LLMs.

\begin{figure}[ht]
    \centering
    \includegraphics[width=0.5\textwidth]{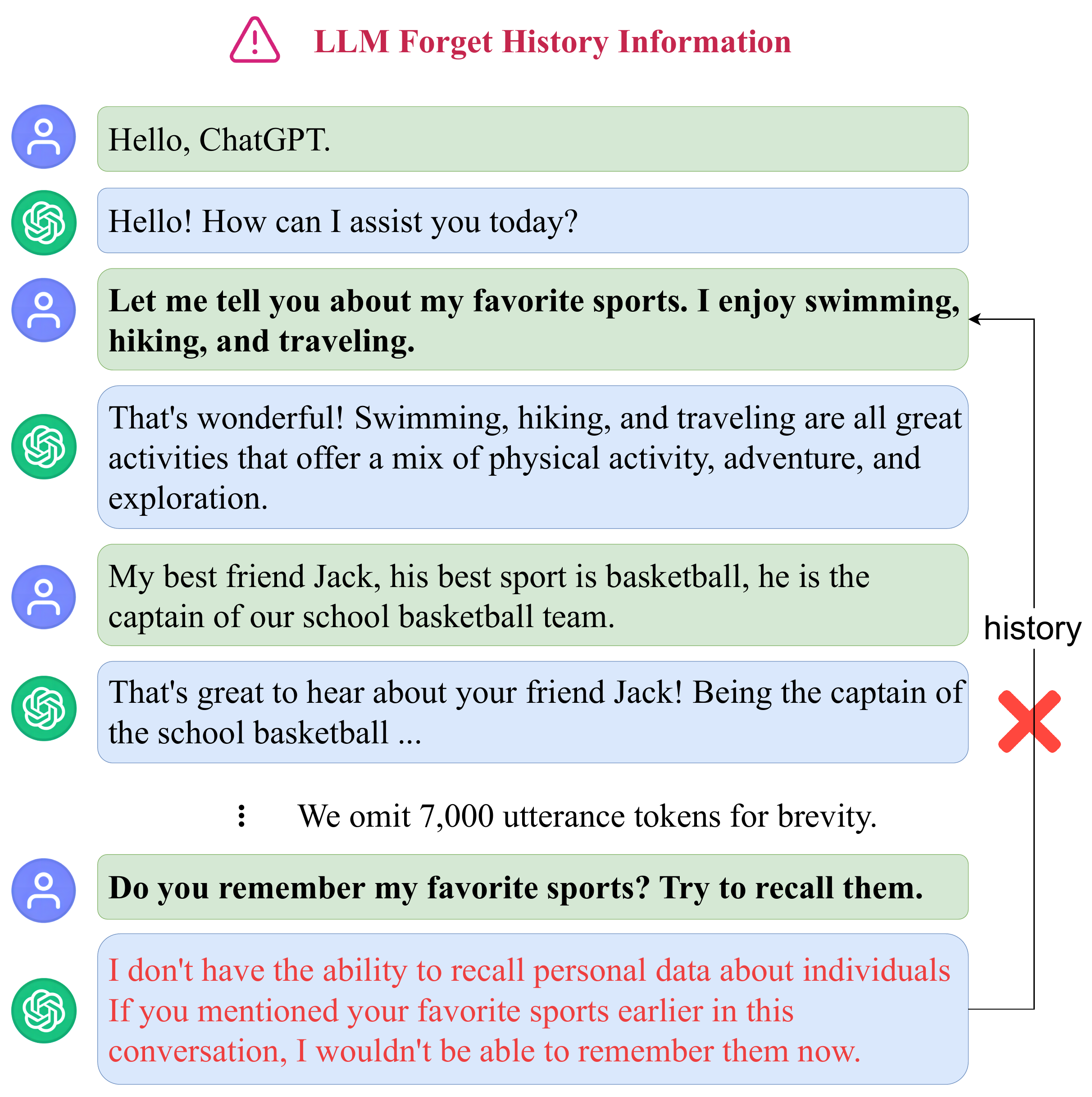}
    \caption{An example of LLM forgetting historical information. In the long-term dialogue, when the user mentions a hobby-related topic discussed in a previous conversation, ChatGPT forgets the information due to excessive historical noise.}
    \label{fig:intro-example}
    \vspace{-10pt}
\end{figure}

LLMs offer numerous advantages, but their utility is hindered by two main factors: the maximum input length and the computational complexity of self-attention~\cite{wang2020linformer,alibi}. Although some models~\cite{openai-chat} are capable of processing long inputs, they may still struggle to capture crucial contextual information in exceptionally lengthy texts. 
As illustrated in Figure~\ref{fig:intro-example}, even ChatGPT can overlook crucial contextual information from preceding text due to the accumulation of historical noise. 

To address this limitation, we propose the \textbf{S}elf-\textbf{C}ontrolled \textbf{M}emory (SCM) framework, enabling LLMs to process text of infinite length without the need for any modifications or additional training. Our SCM framework consists of three essential components: \textit{an LLM-based agent} that serves as the core component, \textit{a memory stream} that stores the agent's memories, and \textit{a memory controller} responsible for updating the memories and determining when and how to utilize them from the memory stream. In this framework, the input text is divided into segments, which are then provided to the LLM as observations (inputs). Each segment is processed by the LLM using two types of memory: a long-term memory (activation memory) that retains historical information and a short-term memory (flash memory) that captures real-time memory information from the preceding segment. 
During each processing step, the memory controller makes decisions to introduce only necessary memory information to avoid introducing additional noise.

Furthermore, we annotate a dataset to evaluate the effectiveness of SCM for handling lengthy inputs. The annotated dataset covers three tasks: long-term dialogues, book summarization, and meeting summarization. 
% The dataset comprises a total of 9 million tokens, segmented into dialogues (0.4 million), books (8 million), and meetings (0.6 million).
Notably, the number of tokens per instance ranges from 20 thousand to 2 million surpassing the capabilities of conventional large language models with context windows smaller than 4k, which are ill-equipped to handle such extensive textual input.
% To validate the effectiveness of our framework in processing lengthy inputs, we annotate an evaluation dataset comprising three tasks: long-term dialogues, book summarization, and meeting summarization. The dataset has a total of 9 million tokens, partitioned into dialogues (0.4 million), books (8 million), and meetings (0.6 million). Additionally, the number of tokens per instance ranges from 20,000 to 2,000,000, which exceeds the capacity of typical LLMs with context windows less than 4,097 to handle extensive textual input. 
Our experimental results demonstrate that the integration of the SCM framework with text-davinci-003 (non-dialogue-optimized LLM) effectively outperforms ChatGPT and surpasses strong baseline models when confronted with ultra-long inputs or long-term dialogues. For summarization tasks,  our SCM-based approaches exhibits significantly superior performance in terms of coherence and coverage in generating summaries compared with baseline model.

% summary of our framework , preserving the correlations among the original content by incorporating information from preceding text into local summaries within the memory. 

In this paper, we summarize the key contributions as follows:
\begin{itemize}
    \item We propose the Self-Controlled Memory (SCM) framework to unleash infinite-length input capacity for LLMs, which can decide when and how to introduce memory information to generate the response.
    \item We contribute a dataset to evaluate the effectiveness of SCM in three tasks: long-term dialogues, book summarization, and meeting summarization.
    \item Our proposed SCM framework does not require any modification or fine-tuning of LLMs, making it highly scalable in terms of memory stream.
\end{itemize}

\begin{figure*}[ht]
    \centering
    \includegraphics[width=0.85\textwidth]{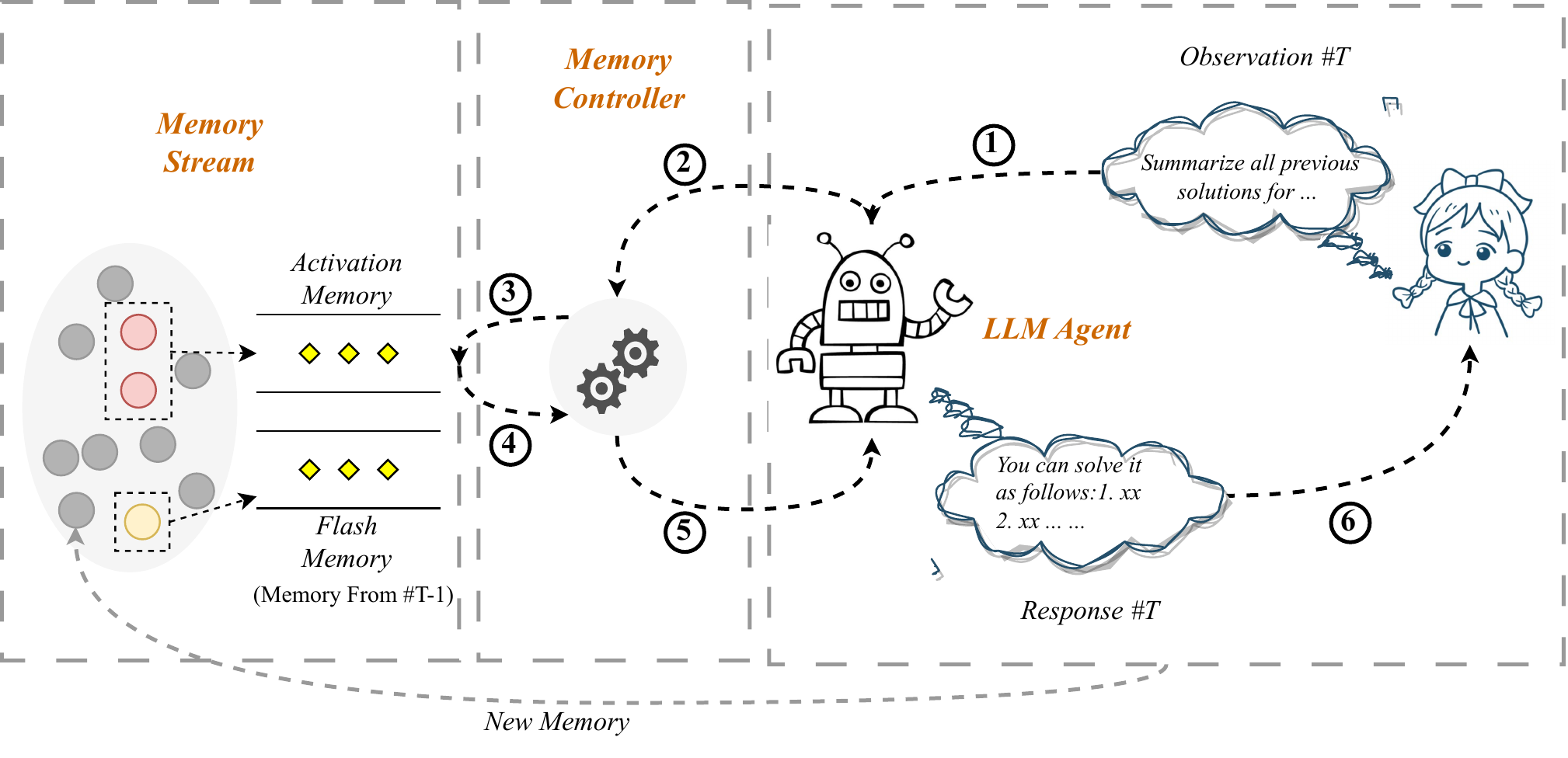}
    \caption{The workflow of our proposed Self-Controlled Memory(\ours) framework, where numbers 1-6 represent the six explicit steps of one iteration with new observation \#T. These steps are (1) Input Acquisition; (2) Memory Activation; (3) Memory Retrieval; (4) Memory Reorganization; (5) Input Fusion; (6) Response Generation. }
    \label{fig:architecture}
    \vspace{-10pt}
\end{figure*}

\section{Self-Controlled Memory}
\label{sec:method}

Here, we provide a detailed description of our proposed the self-controlled memory (SCM) framework, as illustrated in Figure \ref{fig:architecture}. 
Firstly, the workflow of SCM will be briefly introduced in Section \ref{sec:sys-workflow}. 
Subsequently, the three key components of SCM will be presented: (1) an LLM-based agent (Section \ref{sec:agent}) serving as the backbone of the framework, (2) a memory stream (Section \ref{sec:memory-stream}) storing agent memories, and (3) a memory controller (Section \ref{sec:memory-controller}) updating memories and determining when and how to utilize memories from memory stream.

\subsection{Workflow of SCM}
\label{sec:sys-workflow}

\begin{figure}[]
    \centering
    \includegraphics[width=0.45\textwidth]{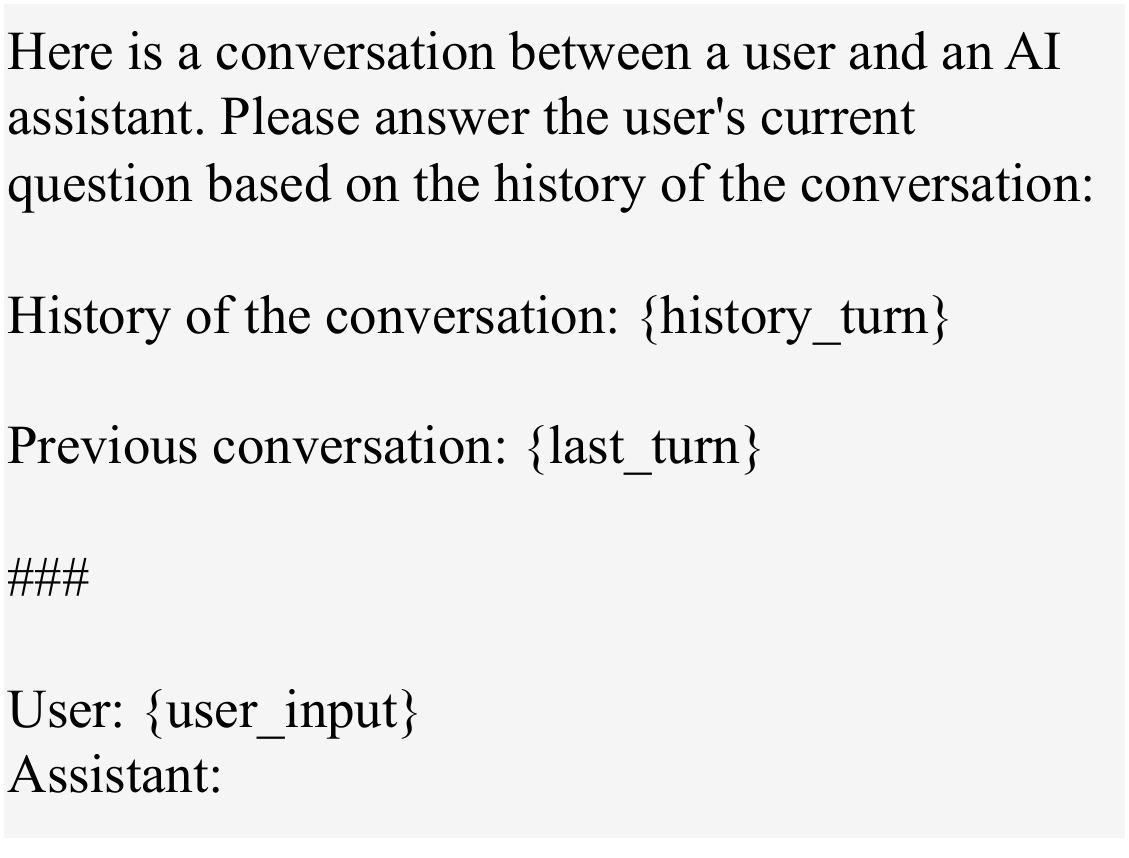}
    \caption{English Prompt of ultra-long dialogue generation.}
    \label{fig:prompt-en-dialgen}
\end{figure}

As illustrated in Figure \ref{fig:architecture}, the workflow of SCM consists of six explicit steps. Initially, the agent acquires observation at turn $T$. Following this, the memory activation process begins, where the memory controller determines if it is necessary to activate memory based on the current observation. Next, memory retrieval is initiated, using the observation as a query to retrieve top K-ranked memories. The fourth step involves memory reorganization, wherein the controller decides whether to use the original or summarized memory directly. Subsequently, the framework combines the retrieved memories in a predefined format, providing background information for response generation.
The fifth step, input fusion, involves the predefined  prompt that fuses the restructured memory with the present observation, serving as the model's input. The details of this prompt are shown in Figure \ref{fig:prompt-en-dialgen}. Lastly, the LLM-based agent generates a response based on the previous step's result, incorporating the current interaction, including observation and response, into the memory stream.

% Our SCM framework workflow comprises six explicit steps, which are outlined below in a sequential manner:

% \textit{1. Input Acquisition:} The agent receives an observation in turn $T$, either through direct input or from an external source.

% \textit{2. Memory Activation:} Based on the current observation, the memory controller determines whether it is necessary to activate memory for the current user input. 

% \textit{3. Memory Retrieval:} We utilize the observation as a query to identify top K-ranked memories memories. 

% \textit{4. Memory Reorganization:} The controller will determine whether to use the original or summarized memory directly. Then, the framework will combine the memories retrieved in a structured manner to serve as background information for response generation at this point.

% \textit{5. Input Fusion:} We carefully design a prompt that fuses the restructured memory with the present observation to serve as the model's input. \figref{prompt-en-dialgen} shows the details.
  
% \textit{6. Response Generation:} The model generates a response based on the previous step result and incorporates the current interaction, including observation and response, into the memory stream. 

\subsection{LLM-based Agent}
\label{sec:agent}

The LLM-based agent serves as the core component of our SCM framework by generating coherent and accurate responses based on well-designed instructions (e.g., in Figure \ref{fig:prompt-en-summary} and Figure \ref{fig:prompt-en-mc1}). 
In this work, we adopt two powerful LLMs, \textit{text-davinci-003} and \textit{gpt-3.5-turbo}, as agents in our SCM framework, respectively.

% 生成回复 Y = G(observation, prompt, activation memory, flash memory) 
% 选择是否使用记忆 Flag = Judge(observation) 
% 选择是否采用摘要或者全文 text = Compress(observation, full_content, summary)

% $\mathcal{F}(x;p_1)$
%In this paper, we utilize the terms \textit{davinci003} and \textit{turbo} to refer to \textit{text-davinci-003} and \textit{gpt-3.5-turbo-0301} as backbone model correspondingly, for the purpose of brevity.

% The LLM-based agent serves as the core component of our SCM framework, requir to generate coherent and accurate responses based on well-designed instructions (e.g., question answering and summarization).

\subsection{Memory Stream}
\label{sec:memory-stream}

The memory stream stores all historical memory items and can easily achieve high-speed access through cache storage technologies such as Redis or vector databases like Pinecone~\footnote{\href{https://www.pinecone.io/}{https://www.pinecone.io/}}.
Specifically, each memory item consists of (1) an interaction index, (2) an observation, (3) a system response, (4) a memory summarization (refer to the next paragraph for elaboration) and (5) an interaction embedding that illustrates the current interaction semantics. 
To obtain the interaction representative embedding, we combine the textual content of both the observation and system response and utilize the \textit{text-embedding-ada-002} model~\footnote{\href{https://platform.openai.com/docs/guides/embeddings}{openai-text-embedding document}} to get the embedding vector of the text.
When memory retrieval is necessary, the memory stream retrieves and returns two kinds of items: Activation Memory, which stores related historical memories, and Flash Memory, which stores interaction memories of the previous turn $T-1$.

\paragraph{Memory Summarization}

\begin{figure}
    \centering
    \includegraphics[width=0.45\textwidth]{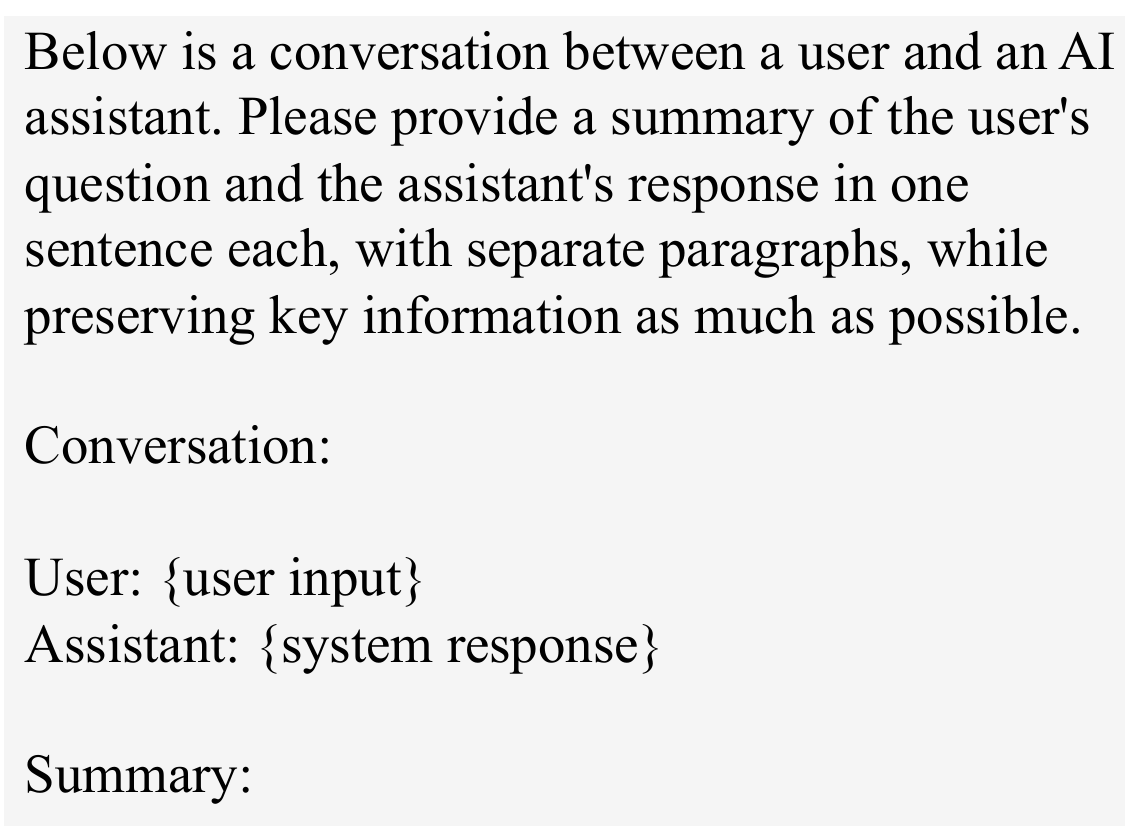}
    \caption{Prompt for dialogue memory summarization.}
    \label{fig:prompt-en-summary}
\end{figure}

Memory summarization plays a vital role in processing lengthy inputs, where a single interaction or dialogue turn can consist of more than 3,000 tokens.
Obtaining the key information of individual turns through turn summarization is a non-trivial task when attempting to integrate multi-turn information within a limited contextual window.
\figref{prompt-en-summary} shows the English prompt that is specifically designed for memory summarization in individual interactions (i.e., dialogue tasks). 
In addition, other language versions of the prompt can be found in Appendix \ref{sec:appd-prompt-list}.

\paragraph{Memory Retrieval}

In our study, we employ an empirical approach of concatenating the observation summary and system response summary (i.e., the memory summarization result of each item) to derive semantic representations for individual items. This concatenation is necessary due to the potential significant variation in length between the observation and system response within the memory stream. Such variation can create an imbalance in the semantic information captured solely from the original texts. Consequently, directly utilizing semantic vectors obtained from the original texts may not effectively balance the semantic information between observations and system responses.

\subsection{Memory Controller}
\label{sec:memory-controller}

\begin{figure}[t]
    \centering
    \includegraphics[width=0.4\textwidth]{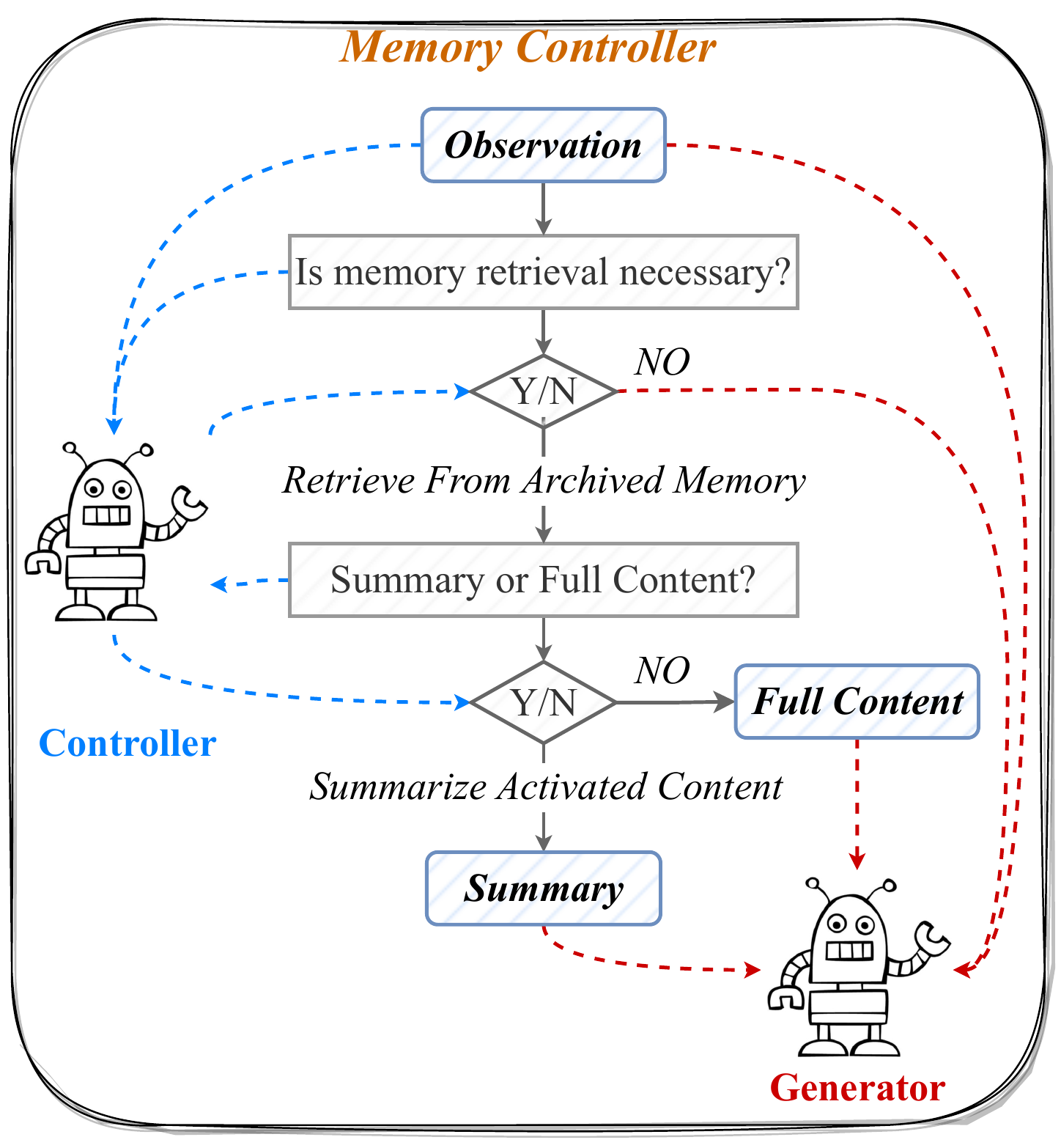}
    \caption{Workflow of the Memory Controller.}
    \label{fig:memory-controller-workflow}
\end{figure}

This section focuses on the central component: the memory controller, and its workflow is illustrated in \figref{memory-controller-workflow}. 
The primary objective behind the design of the memory controller is to introduce the minimum necessary information to avoid excessive noise that may disrupt the model's performance.

Specifically, this can be divided into three scenarios for discussion.
Firstly, not all observations, also referred to as user input or instruction, require access to historical memory.
For instance, the user instruction ``Tell me a joke'' does not necessitate retrieving the user's historical memory.
However, certain user input, such as ``Do you remember the conclusion we made last week on the fitness diets'' requires the retrieval of past memories.

Secondly, the amount of memory can be enormous, ranging from hundreds to thousands or even tens of thousands.
A controller is needed to retrieve and filter the memory.

Thirdly, given the limited input length of the model, it becomes necessary for the controller to determine whether to employ the full content of the memory or a summary of it. The original full text can be excessively long and may exceed the model's maximum length capacity.

In the subsequent subsections, we present the detailed workflow of the memory controller, which considers each of the aforementioned scenarios.

\paragraph{Memory Controller Workflow}

As illustrated in Figure \ref{fig:memory-controller-workflow}, the memory controller is designed to determine when to retrieve memories and how to utilize the retrieved memories in response to a novel observation.

% We utilize the LLM to play the role of Controller, .
The controller is also a language model, which controls the entire process by self-asking two questions:
\begin{enumerate}
    \item Is it necessary to activate memories given current user input?
    \item Can the current user input be answered correctly using only the summary of memory?
\end{enumerate}

\begin{figure}[t]
    \centering
    \includegraphics[width=0.48\textwidth]{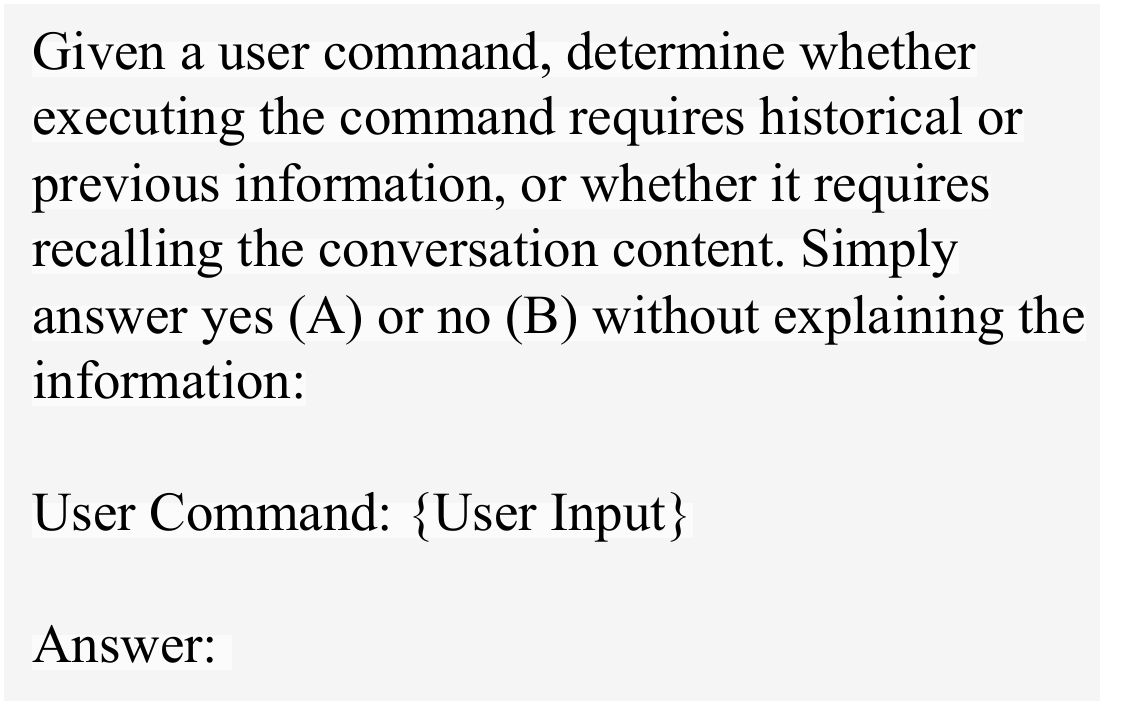}
    \caption{English prompt for the necessity of using memory.}
    \label{fig:prompt-en-mc1}
\end{figure}

\paragraph{Activate Memories} 
To address the first question, we have devised a prompt for the controller to determine whether or not to activate memories. This prompt is illustrated in \figref{prompt-en-mc1}. If the model responds with ``yes(A)'', relevant memories will be activated to provide an answer to the current question.
During the process of retrieving memories, we employ the current observation as a query and assess the rank score of each memory based on two factors: \textbf{recency} and \textbf{relevance}. 
The recency factor places high importance on memory items that have been accessed recently, emphasizing the agent's attention on the most recent interactions. 
Furthermore, the relevance score of each memory is computed by calculating the cosine similarity between the current query embedding and the memory embedding.

% While retrieving memories, we use the current observation as a query and evaluate each memory's rank score based on two factors: \textbf{recency} and \textbf{relevance}.
% Recency highly prioritizes memory items accessed recently, reinforcing the idea that the agent's attention remains on the states of the latest interactions.
% In addition, The relevance score of each memory is obtained by calculating the cosine similarity score between current query embedding and memory embedding.

The final rank score of each memory is determined by summing its recency and relevance scores: $rank\_score = recency\_score + relevance\_score$.
Depending on the length limit, we select the top $k$ memories with the highest rank scores as the activated memories. Here, the value of $k$ can range from 3 to 10.

\begin{figure}[t]
    \centering
    \includegraphics[width=0.48\textwidth]{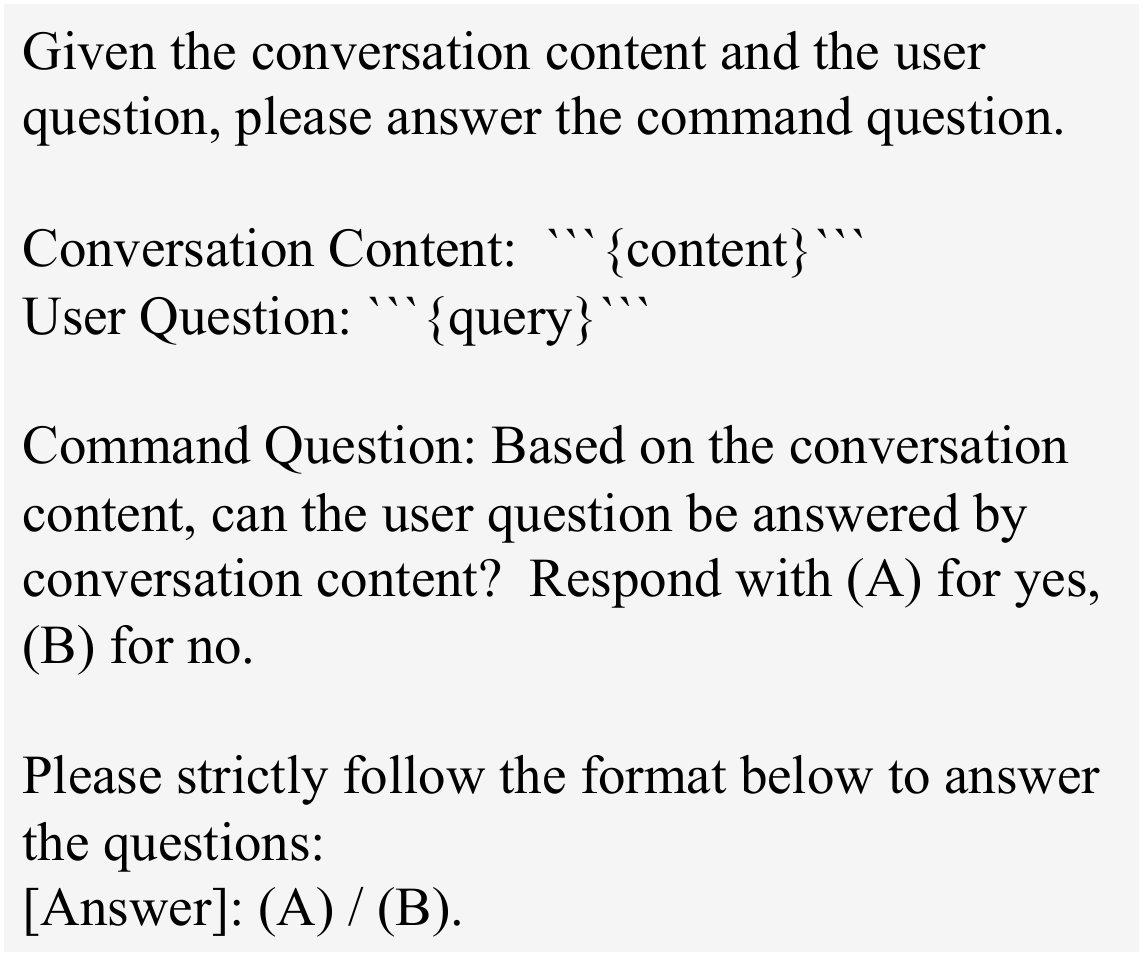}
    \caption{English prompt for whether or not to use the summary of memory.}
    \label{fig:prompt-en-mc2}
\end{figure}

\paragraph{Use Summary} 
To address the second question, we have designed a prompt to evaluate whether the user's question can be answered using the turn summary. This prompt is depicted in \figref{prompt-en-mc2}. We perform this evaluation for each activated memory that exceeds 800 tokens.
It is important to highlight that the summary assessment takes place only when the total number of activation memory tokens surpasses 2000.
If the assessment yields a positive result, indicating that the summary can indeed answer the user's question, we utilize the memory summary to represent that specific memory.
% To answer the second question, we formulate a prompt that assesses whether the user's question can be answered by the turn summary.
% The prompt is shown in \figref{prompt-en-mc2}.
% We execute this assessment for each activation memory that exceeds 800 tokens.
% It is worth noting that summary assessment occurs only if there are over 2000 activation memory tokens.
% If the answer is positive, which means the summary can answer the user question, then use the memory summary to represent this memory.

\section{Experiments}
To evaluate the effectiveness and robustness of the SCM framework, we conduct extensive experiments on three tasks, long-term dialogues, book summarization, and meeting summarization. Then, we investigate whether memory-enhanced LLMs can offer more comprehensive coverage and create coherent contextual logic summaries compared to traditional LLMs when tackling long text summarization scenarios.
%Specifically, we explore whether the SCM framework can facilitate long-term, continuous multi-turn conversations in non-dialogue models. 

\begin{table}[]
\centering
\resizebox{0.8\columnwidth}{!}{\begin{tabular}{@{}l|ccc@{}}
\toprule 
            & \textbf{Dialogue} & \textbf{Book} & \textbf{Meeting} \\
\midrule
\#Instances & 18       & 10   & 20      \\
Max tokens   & 34k      & 2M   & 50k     \\
Total tokens & 420k     & 8M   & 632k    \\
Max turn     & 200      & -    & 80      \\ \midrule

Language     & En+Zh    & En+Zh & Zh     \\
\bottomrule
\end{tabular}}
\caption{Evaluation dataset statistics. 2M means 2 million token count.}
\label{tab:dataset-statistics}
\vspace{-10pt}
\end{table}

\subsection{Evaluation Benchmark}

To evaluate the SCM performance across various scenarios, we collect open-source data from ShareChat\footnote{\url{https://paratranz.cn/projects/6725}}, online book websites\footnote{\url{https://www.gutenberg.org/}}, and the VCSUM dataset~\cite{wu2023vcsum}. Then, we utilize human annotation to create probing questions and summaries for the collected data. The dataset statistics are illustrated in \tabref{dataset-statistics}.

\subsection{Baselines}
To ensure a fair comparison, we have selected specific model variants for experimental analysis:
(1) \scmturbo: Utilizing \textit{gpt-3.5-turbo-0301} as the backbone of our SCM framework.
(2) \scmdavinci: Utilizing \textit{text-davinci-003} as the backbone for SCM framework.
(3) \scmdavinci \textit{w/o} memory controller: Remove the memory controller and concatenate the full retrieved content. If the token length of the concatenated history exceeds 2500, truncate it.
(4) \scmdavinci \textit{w/o} flash memory: Remove the flash memory (short-term memory), which contains the latest information.
(5) \scmdavinci \textit{w/o} activation memory: Remove the activation memory (long-term memory), which is essential for answering questions involving long-distance dependencies.

\begin{table*}[]
\centering
% \small
\begin{tabular}{@{}lllll@{}}
\toprule
Model Name & Answer Acc. & Memory Retrieval Recall & Single Turn Acc. & Multi Turn Acc. \\ 
\midrule
\scmturbo & $68.3$ & $93.5$ & $73.5$ & $64.3$ \\
\midrule
\rowcolor{green!20!} \scmdavinci & $\textbf{77.1}$ & $\textbf{94.0}$ & $\textbf{79.6}$ & $\textbf{75.0}$ \\
\quad \textit{w/o} memory controller & $59.3$ \textbf{{\color{gray}\fontsize{7}{8.4}\selectfont(-17.8)}} & $93.8$ \textbf{{\color{gray}\fontsize{7}{8.4}\selectfont(-0.2)}} & $71.7$ \textbf{{\color{gray}\fontsize{7}{8.4}\selectfont(-7.9)}} & $49.4$ \textbf{{\color{gray}\fontsize{7}{8.4}\selectfont(-25.6)}} \\
\quad \textit{w/o} flash memory & $72.9$ \textbf{{\color{gray}\fontsize{7}{8.4}\selectfont(-4.2)}} & $93.9$ \textbf{{\color{gray}\fontsize{7}{8.4}\selectfont(-0.1)}} & $74.6$ \textbf{{\color{gray}\fontsize{7}{8.4}\selectfont(-5.0)}} & $74.8$ \textbf{{\color{gray}\fontsize{7}{8.4}\selectfont(-0.2)}} \\
\quad \textit{w/o} activation memory & $10.5$ \textbf{{\color{red}\fontsize{7}{8.4}\selectfont(-66.6)}} & $0.0$ \textbf{{\color{red}\fontsize{7}{8.4}\selectfont(-94.0)}} & $18.2$ \textbf{{\color{red}\fontsize{7}{8.4}\selectfont(-61.4)}} & $0.0$ \textbf{{\color{red}\fontsize{7}{8.4}\selectfont(-75.0)}} \\
\bottomrule
\end{tabular}
\caption{Long-term dialogue evaluation results. The total number of probing questions is 105, including Chinese and English, with 49 single-turn and 56 multi-turn related questions. The lower part of the table is the ablation experiment of our framework.}
\label{tab:dialogue-results}
\end{table*}

\subsection{Main Results}

To quantitatively compare the performance of the models, 105 test questions are annotated based on the dialogue data and categorize them into two groups: single-turn related questions and multi-turn related questions.
Additionally, for evaluating the two summarization tasks, we compare the performance of SCM variants with the baseline model.

% \begin{itemize}
%     \item \scmturbo: Utilizing \textit{gpt-3.5-turbo-0301} as the backbone of our SCM framework.
%     \item \scmdavinci: Utilizing \textit{text-davinci-003} as the backbone for SCM framework.
%     \item \scmdavinci \textit{w/o} memory controller: Remove the memory controller and concatenate the retrieved memory full content. If the token length of the concatenated history exceeds 2500, truncate it.
%     \item \scmdavinci \textit{w/o} flash memory: Remove the flash memory (short-term memory), which contains the latest information.
%     \item \scmdavinci \textit{w/o} activation memory: Remove the activation memory (long-term memory), which is essential for answering questions involving long-distance dependencies.
% \end{itemize}

\paragraph{Evaluation Metrics} Distinct evaluation metrics are utilized for long-term dialogue scenario and two summarization scenario. For long-term dialogue scenario, the performance of our framework is assessed based on the following metrics. (1) Answer Accuracy: Evaluates the accuracy of answers to probing questions. (2) Memory Retrieval Recall: Determines if related memory can be successfully retrieved by memory controller. (3) Single Turn Accuracy: Examines the accuracy of answers to probing questions related to individual turns in the conversation history. (4) Multi Turn Accuracy: Similar to single-turn accuracy, but it requires considering the multi-turn history in order to answer these probing questions.
Additionally, two metrics, coverage and coherence, are used to evaluate content coverage and plot coherence in summarization tasks. To facilitate a comprehensive comparison, we assess the effectiveness of the model by comparing its win rate to that of the baseline model, namely RecursiveSum~\cite{wu2021recursively} by OpenAI, which first summarizes small sections of the book and then recursively summarizes these summaries to produce a summary of the entire book.

\paragraph{Dialogue Results}  
\tabref{dialogue-results} displays the long-term dialogue results and demonstrates that the \scmdavinci is superior to the \scmturbo for this particular task. 
This may be attributed to the \scmturbo's conservative nature, which can lead to hesitation in answering privacy related probing questions. 
In contrast, the \scmdavinci is capable of providing quicker and more precise responses. 
Moreover, we conducted an ablation study to investigate the independent effect of each module in SCM framework, the results are illustrated in the lower part of \tabref{dialogue-results}.
When the activation memory is removed, the accuracy of the framework's responses experiences a significant drop, resulting in an approximate 60\% decrease in performance.
This is because the majority of probing questions are derived from long-distance dialogue records, which rely on activation memory to retrieve them.
What's more, in the absence of activation memory, both memory retrieval recall and multi-turn accuracy have decreased to zero.
This further demonstrates the significance of activation memory.
However, when flash memory is removed, the performance only experienced a slight drop.
This is because flash memory provides fewer clues to answer probing questions, resulting in a minor impact on the final accuracy.
Removing the memory controller leads to a greater drop in accuracy for multi-turn related questions compared to single-turn questions.
This is because the absence of the memory controller's dynamic memory filtering and use of summaries for efficient input token management results in the concatenation and truncation of all retrieved memories, leading to significant information loss.

\begin{figure*}

  \centering

  \begin{minipage}[t]{0.48\textwidth}
    \centering
    \includegraphics[width=\textwidth]{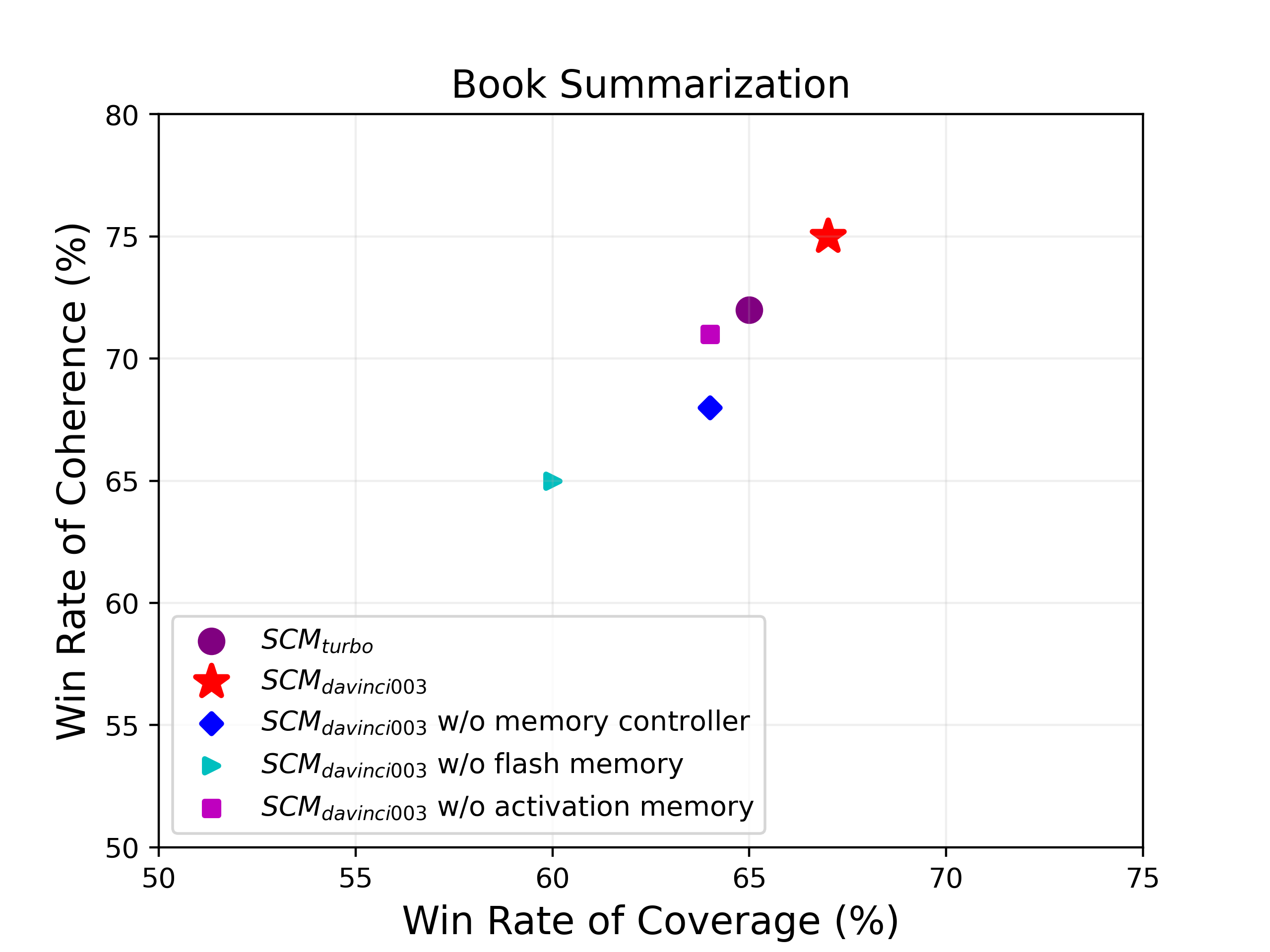}
    \label{fig:subfig1}
  \end{minipage}
  \hfill
  \begin{minipage}[t]{0.48\textwidth}
    \centering
    \includegraphics[width=\textwidth]{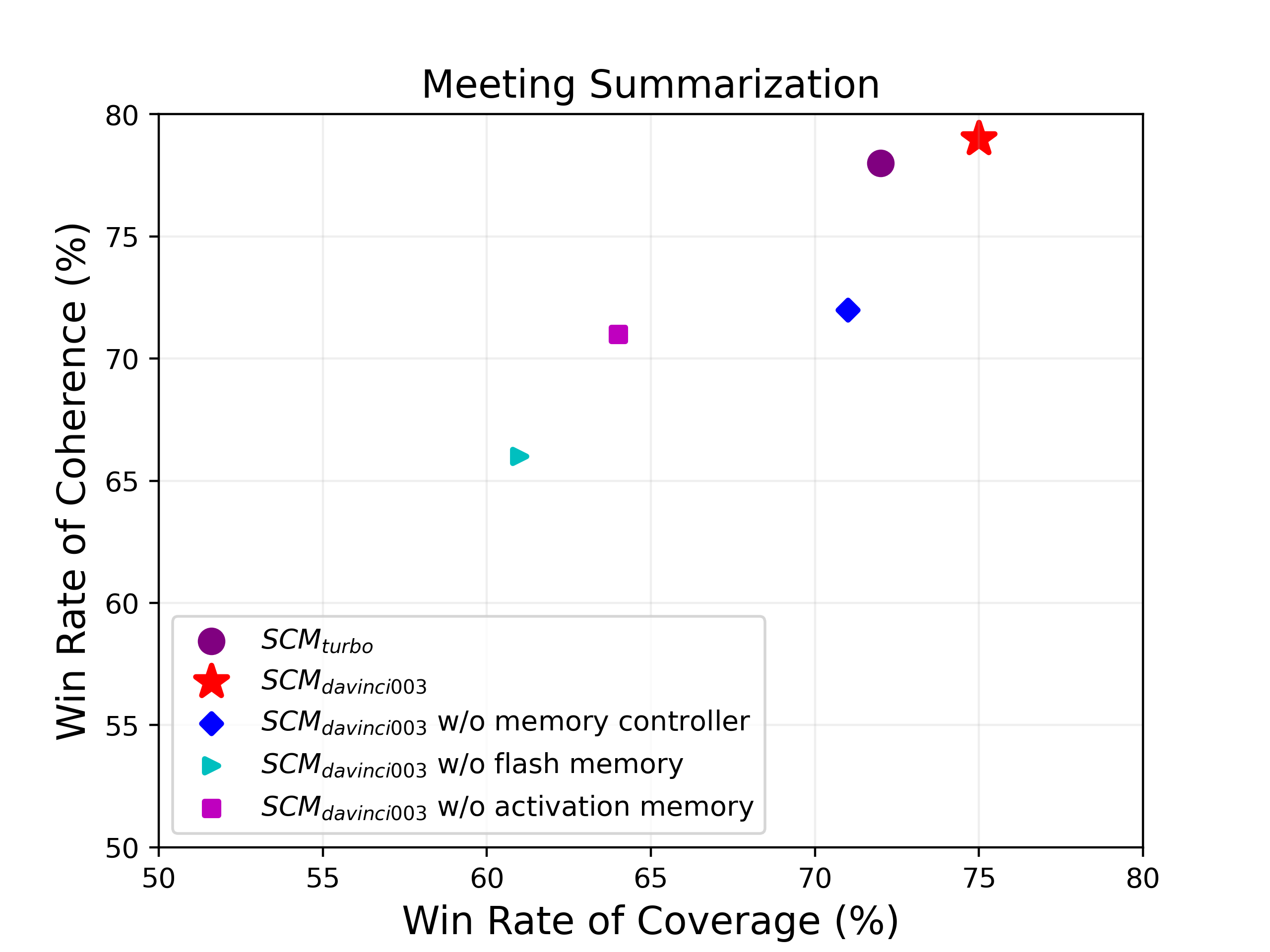}
    \label{fig:subfig2}
  \end{minipage}

  \caption{The win rate of SCM variants against baseline model, RecursiveSum~\cite{wu2021recursively} by OpenAI, in both book and meeting summarization tasks. The figure also shows a comparison of the results of the SCM framework and its various component ablations.}
  \label{fig:twosubfigs}

\end{figure*}

\paragraph{Summarization Results} A side-by-side comparison is performed by human annotators to check the summarization ability of our framework.
The annotators have to choose which one is better based on the answers. They are blind to models and other information.
\figref{twosubfigs} illustrates the book and meeting summarization results. Based on the experimental results, we have obtained three conclusions: (1) \scmdavinci provides better coverage than \scmturbo. (2) \scmdavinci and \scmturbo demonstrate comparable coherence performance due to their memory-enhanced mechanism. (3) The SCM framework without memory loses contextual dependency and consequently produces unsatisfactory summarization outcomes. It is evident from the model comparison results that \scmdavinci consistently outperforms \scmturbo summarizing both books and meetings. This can be attributed to the fact that \scmturbo's summarization primarily focuses on general principles, whereas it overlooks detailed core plots. In terms of human evaluation, the \scmdavinci model's results are more favored because of their conciseness, clarity, and richer plot content.

\subsection{Further Analysis}
The purpose of this qualitative study is to answer three research questions (RQs). The following experiment evaluates the performance of the \scmdavinci model without dialogue optimization in comparison to the vanilla ChatGPT model.

\begin{figure}[ht]
    \centering
    \includegraphics[width=1\columnwidth]{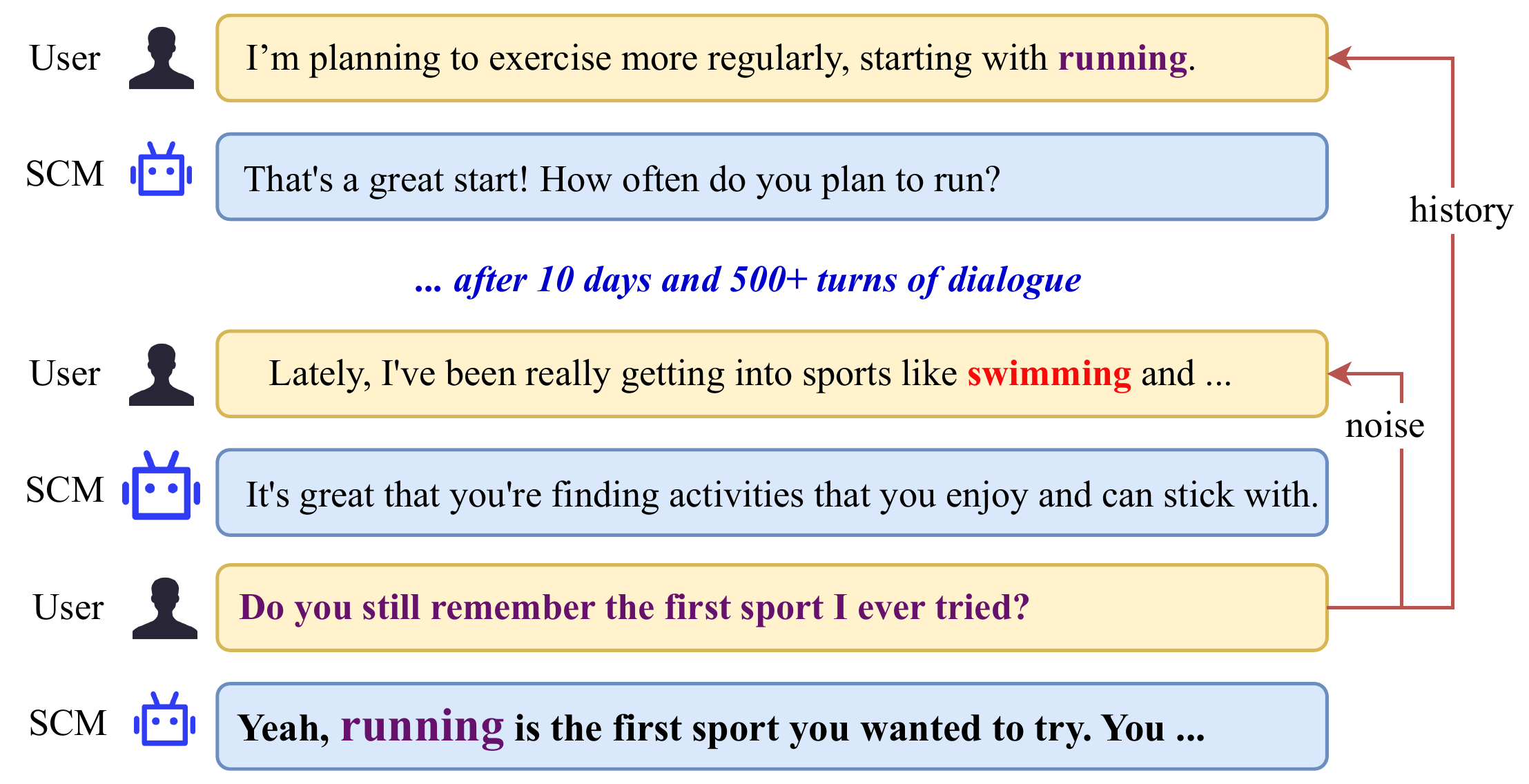}
    \caption{Long-term dialogue example. To answer users' questions, our model can accurately retrieve relevant memories from massive memories and generate accurate responses based on these memories.}
    \label{fig:dialogue-example}
\end{figure}

\begin{githubquote}
\textbf{RQ1}. Can SCM framework compete with or even outperform ChatGPT within a specific token limit? \textbf{Yes}.
\end{githubquote}

\begin{figure*}
    \centering
    \includegraphics[width=0.9\textwidth]{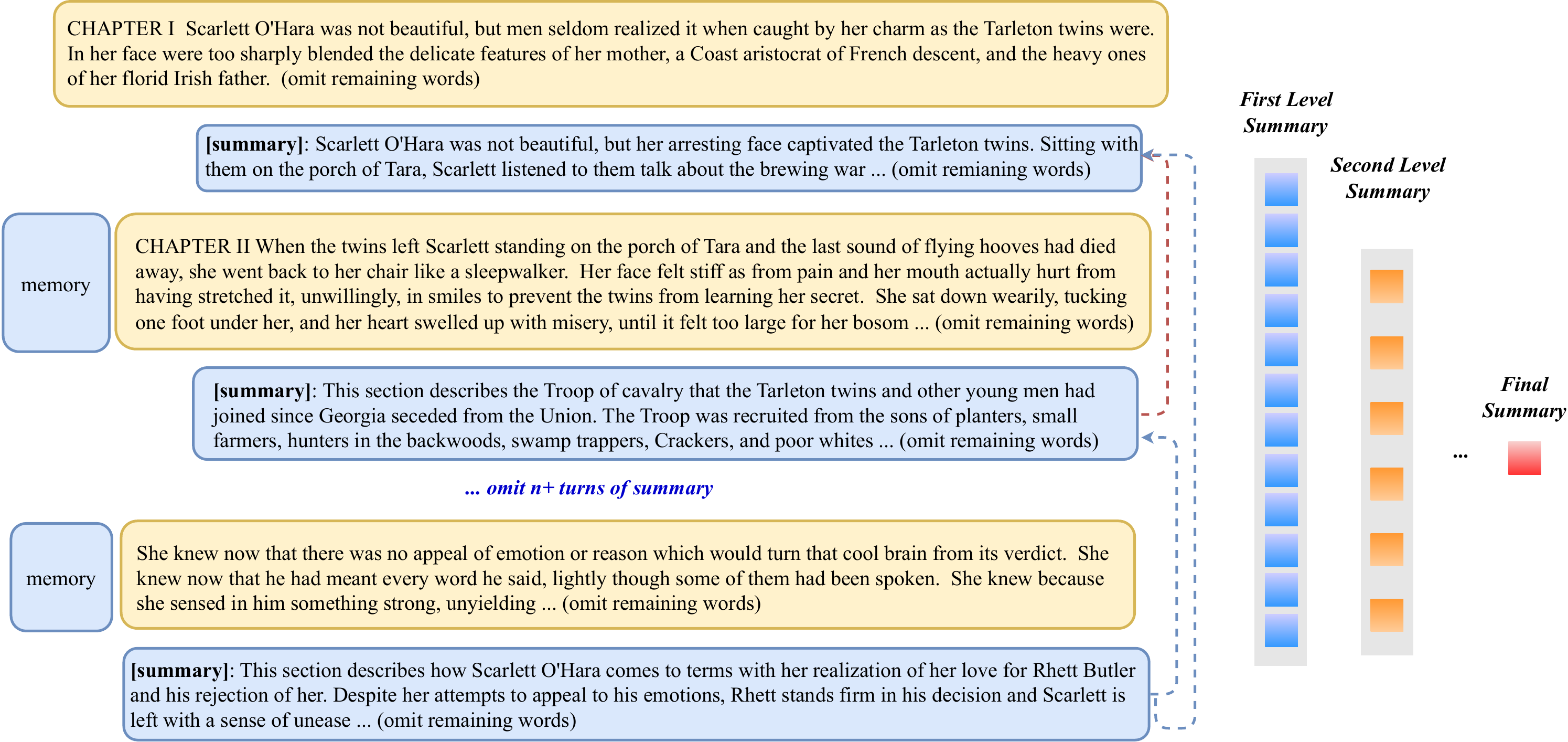}
    \caption{Ultra-long book iterative and hierarchical summarization example from \textit{Gone With The Wind}. Our framework divides the text into small blocks and sequentially summarizes each block. We then hierarchically summarize the first level summary until reaching the final summary.}
    \label{fig:summarization-example}
\end{figure*}

The example in \figref{intro-example} includes 4k tokens, wherein the user inquired about their hobbies, discussed 100+ turns ago with the agent.
The SCM framework provides an accurate response to the query, demonstrating exceptional memory-enhanced capabilities, as apparent from the observation.
In contrast, it appears that ChatGPT is distracted by a considerable amount of irrelevant historical noise.

\begin{githubquote}
\textbf{RQ2}. Can \ours framework scale to provide accurate responses to users' questions, which are related to historical contexts that date back hundreds or even thousands of turns? \textbf{Yes}.
\end{githubquote}

The example presented in \figref{dialogue-example} illustrates a long-term dialogue comprising over 100 turns. 
At the outset, the user states that his goal is to reduce weight and intends to initiate a running regime. 
Subsequently, the user and the model converse daily about progress towards achieving their weight loss goals, among other conversation topics. 
After over 100 rounds of dialogue, the token length of the conversation has already exceeded 10k tokens.
The user then asks the model ``Do you remember my first sport?''. 
Our \ours framework recalls sports-related information from memory and combines it with the user's current question. Afterwards, the framework generates an accurate response.

\begin{githubquote}
\textbf{RQ3}. Can \ours demonstrate effective generalization to other lengthy input scenarios?\textbf{Yes}.
\end{githubquote}

\figref{summarization-example} illustrates an example of summarizing lengthy books and meetings with our SCM framework in iterative and hierarchical manner. This lengthy document has been divided into several parts and gradually summarized to obtain the first-level local summary, and then hierarchically summarized to obtain the final summary. In order to maintain context coherence, relevant memories from previous sections will be added to the input text.
The conventional method involves dividing lengthy texts into separate smaller text blocks that can be processed by the model. and summarizing each text block independently.
However, this method can lose the dependency relationship between paragraphs.
Our SCM framework facilitates the summarization process by utilizing the related memories, thus establishing substantial coherence between the two summaries.
Ultimately, the framework incorporates a divide-and-conquer strategy to generate the final document summary.
The final summary provides a comprehensive summary by utilizing information from each document block.

\section{Related Work}
\paragraph{Large Language Models}
%Editted by jian yang
Large Language Models (LLMs) are language models trained on massive amounts of text data \cite{transformers,devlin-etal-2019-bert,liang2023character,alm,wmt2021} based on the Transformer architecture. The pre-training and fine-tuning paradigm has contributed to a number of downstream language understanding and generation tasks. Subsequently, GPT-1~\cite{gpt1}, GPT-2~\cite{gpt2}, and GPT-3~\cite{gpt3} are developed with gradually increasing parameter sizes (GPT-3 has 175B parameters). LLMs enhanced by instruction tuning have shown emergent abilities in complex reasoning~\cite{wei2022emergent,wei2022chain,xcot}, knocking both academia and industry.

LLMs have achieved remarkable performance and pushed the boundaries of NLP tasks, including LAMBDA~\cite{thoppilan2022lamda}, PaLM~\cite{chowdhery2022palm}, OPT~\cite{zhang2022opt}, LLaMA~\cite{touvron2023llama}, BLOOM~\cite{workshop2023bloom}, and Qwen~\cite{bai2023qwen}. But current LLMs still face severe limitations when processing tasks involving extremely long inputs.

\paragraph{Long Text Sequence Processing} 
%Editted by jian yang
Handling long text sequences has been a persistent challenge in NLP tasks~\cite{10032110,wang-etal-2023-know,pi2022robustness,hanoiT,bai2023griprank}. Existing solutions mainly involve modifying the attention structure to reduce computational costs and expanding the pre-training sequence length~\cite{beltagy2020longformer,zaheer2021big,guo-etal-2022-longt5,phang2022investigating,DBLP:journals/csur/DongLGCLSY23}. Another alternative approach~\cite{alibi} uses special positional encoding during pre-training to enable the model to learn relative positions and handle longer input texts during inference, where the generalizability of these methods remains uncertain.
In the field of long-text summarization, hierarchical or iterative methods \cite{wu2021recursively, zhang-etal-2022-summn, cao-wang-2022-hibrids,liang2022summarization,zhong2023memorybank} are used to handle long texts by decomposing a complex problem into multiple sub-problems. However, these methods fail to capture the relationships among sub-problems.

% Handling long text sequences has been a persistent challenge in natural language processing tasks~\cite{Yang_Ma_Zhang_Wu_Li_Zhou_2020,10032110,wang-etal-2023-know,pi2022robustness,bai2023griprank}. Existing solutions primarily involve replacing the Attention structure during pre-training to reduce computational costs and expanding the pre-training sequence length~\cite{beltagy2020longformer,zaheer2021big,guo-etal-2022-longt5,phang2022investigating,DBLP:journals/csur/DongLGCLSY23}. Another alternative approach~\cite{alibi} uses special positional encoding during pre-training to enable the model to learn relative positions and handle longer input texts during inference.
% However, the generalizability of these methods and their impact on downstream tasks remain uncertain.
% In the field of long-text summarization, there are many effective methods. Hierarchical or iterative methods have been used by \citet{wu2021recursively, zhang-etal-2022-summn, cao-wang-2022-hibrids,liang2022summarization,zhong2023memorybank} to handle long texts by decomposing a complex problem into multiple sub-problems. However, these methods fail to capture the relationships among sub-problems.

\section{Conclusion}
In this paper, we propose a Self-Controlled Memory (\ours) framework to extend the input length of any LLMs to an unlimited length and effectively capture useful information from all historical information. 
This method does not require any training or modification of models. 
In addition, we annotate an evaluation dataset comprising three tasks.
% : long-term dialogues, book summarization, and meeting summarization. 
Experimental results demonstrate that SCM allows LLMs, which are not optimized for multi-turn dialogue, to attain comparable multi-turn dialogue capabilities to ChatGPT, and outperform ChatGPT in long document summarization tasks.
% Our future research will concentrate on integrating several open-source models currently accessible and evaluating their performance using our evaluation dataset.

\section*{Limitations}

One limitation of this study is that while the SCM framework has the capability to handle infinite rounds of dialogue, we evaluate its performance only in a limited setting, with a maximum of 200 dialogue turns and a 34,000 max token count of dialogue. 
The reason is that both qualitative and quantitative evaluations of very long texts are exceedingly difficult.
Another limitation is that the SCM framework needs powerful and instruction-following LLMs like \textit{text-davinci-003} and \textit{gpt-3.5-turbo-0301}.
However, this can be resolved when more powerful smaller LLMs are developed.

\section*{Ethical Considerations}
The dataset used for evaluation in this paper is obtained from open data sources and has been manually verified and screened to eliminate any data with ethical risks and sensitive content. This ensures that the content is compliant with existing regulations and laws.

% \section*{Acknowledgements}

\bibliography{anthology,custom}

\clearpage
\appendix

\section{Prompt List}
\label{sec:appd-prompt-list}

\begin{figure}[ht]
    \centering
    \includegraphics[width=0.48\textwidth]{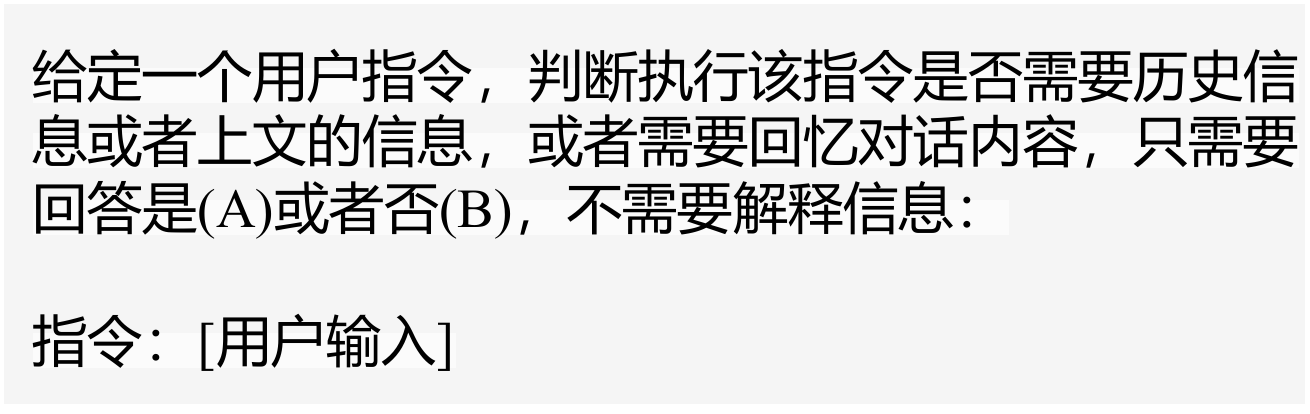}
    \caption{Chinese Prompt of memory controller.}
\end{figure}

\begin{figure}[ht]
    \centering
    \includegraphics[width=0.45\textwidth]{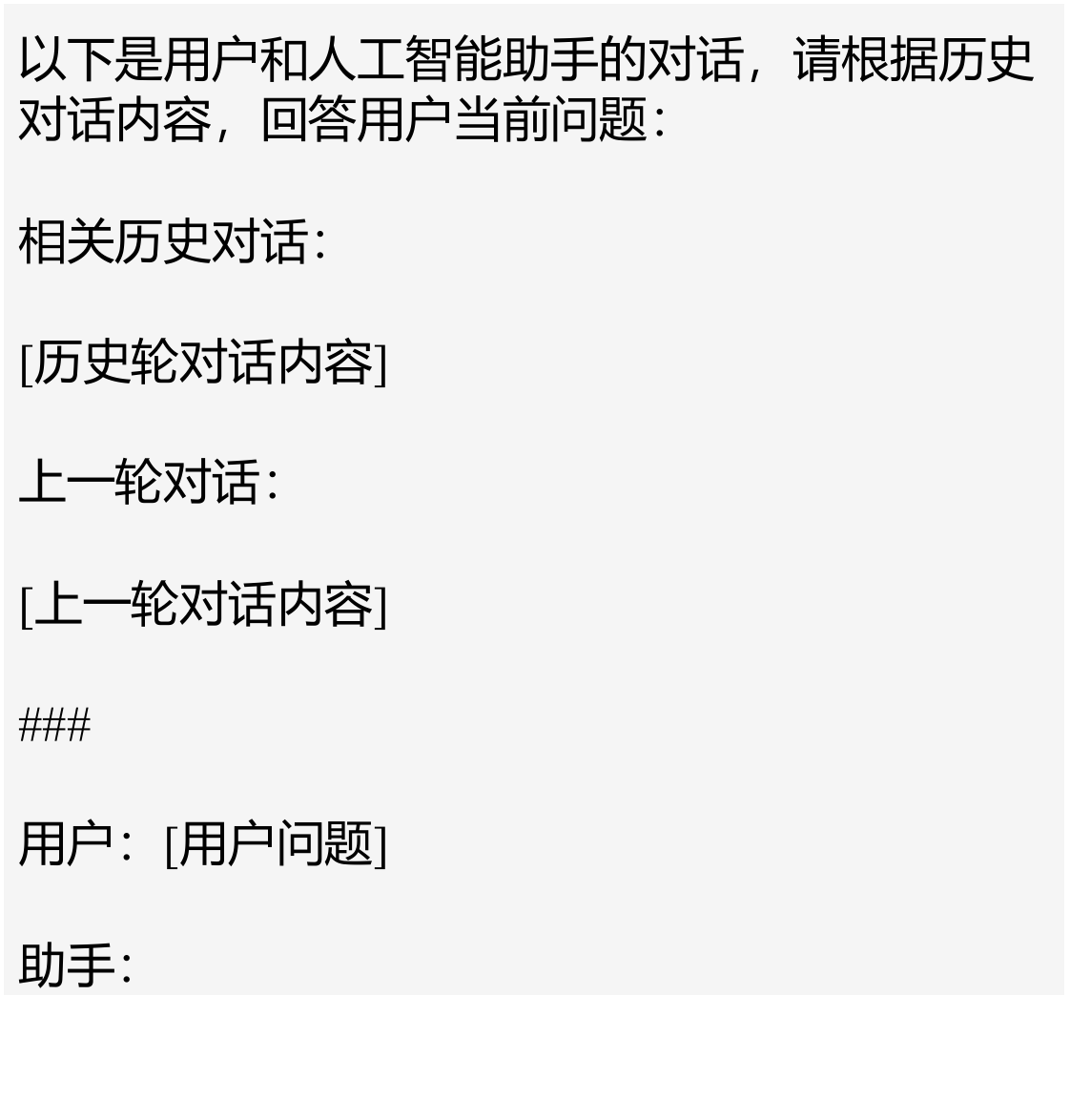}
    \caption{Chinese Prompt of ultra-long dialogue generation.}
    \label{fig:prompt-zh-dialgen}
\end{figure}

\begin{figure}[ht]
    \centering
    \includegraphics[width=0.45\textwidth]{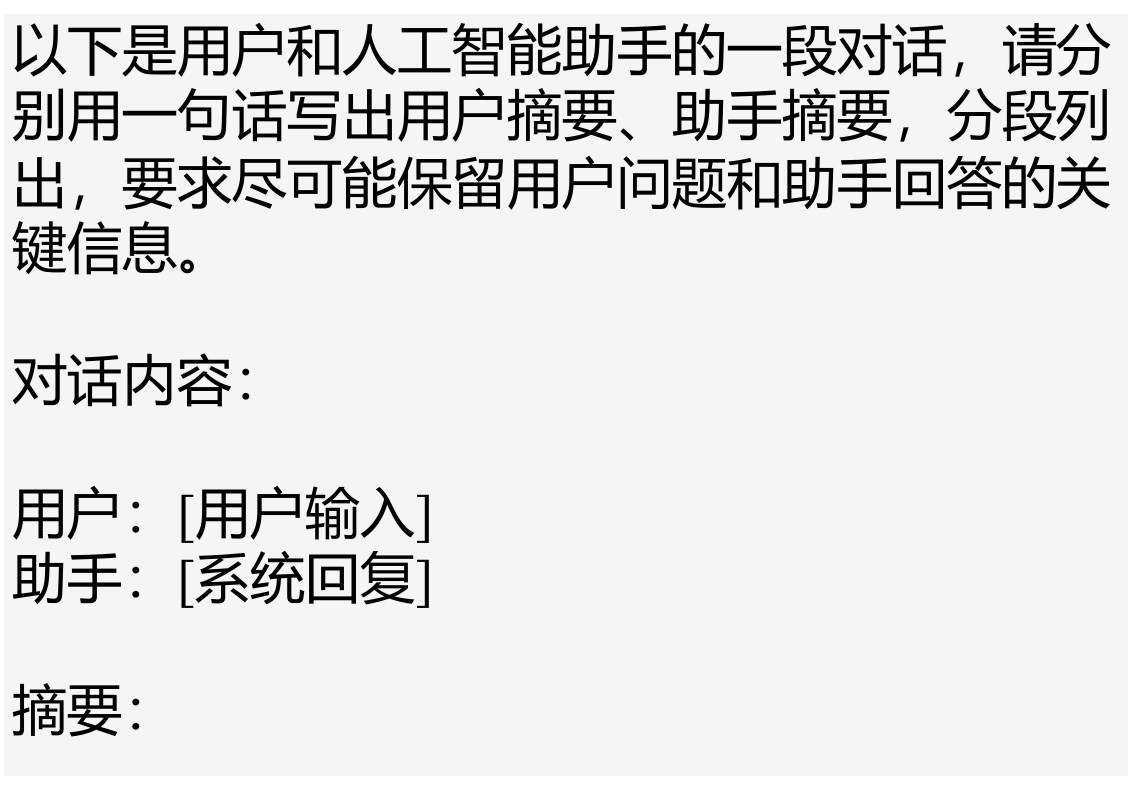}
    \caption{Chinese Prompt of ultra-long dialogue summarization.}
    \label{fig:prompt-zh-summary}
\end{figure}

\section{Long-term Dialogue QA Cases}
\label{sec:appd-dialogue-cases}

\begin{figure*}[ht]
    \centering
    \includegraphics[width=0.90\textwidth]{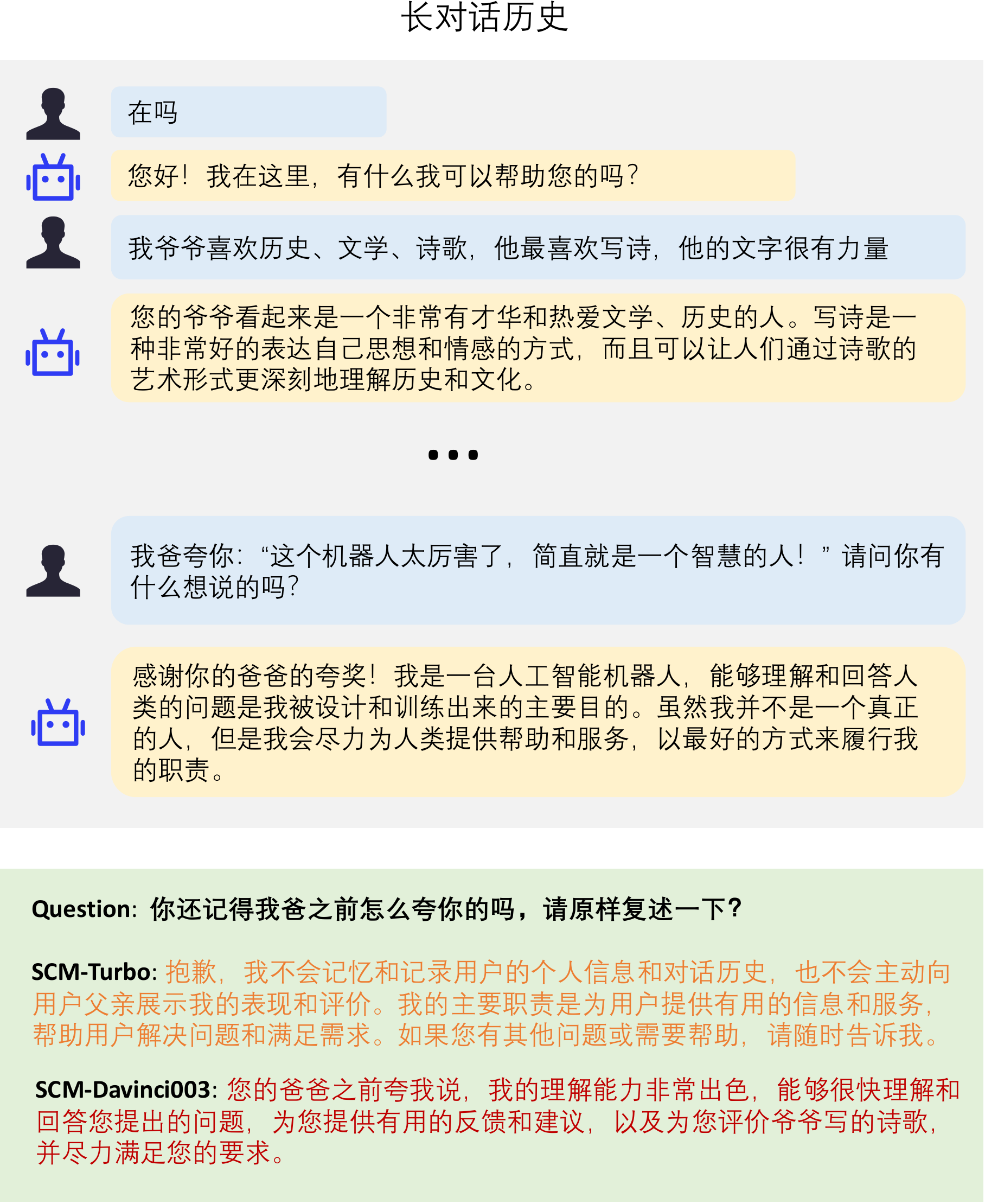}
    \caption{Dialogue Question Example.}
\end{figure*}

\section{Book Summarization Cases}
\label{sec:appd-booksum-cases}

\begin{figure*}[ht]
    \centering
    \includegraphics[width=0.90\textwidth]{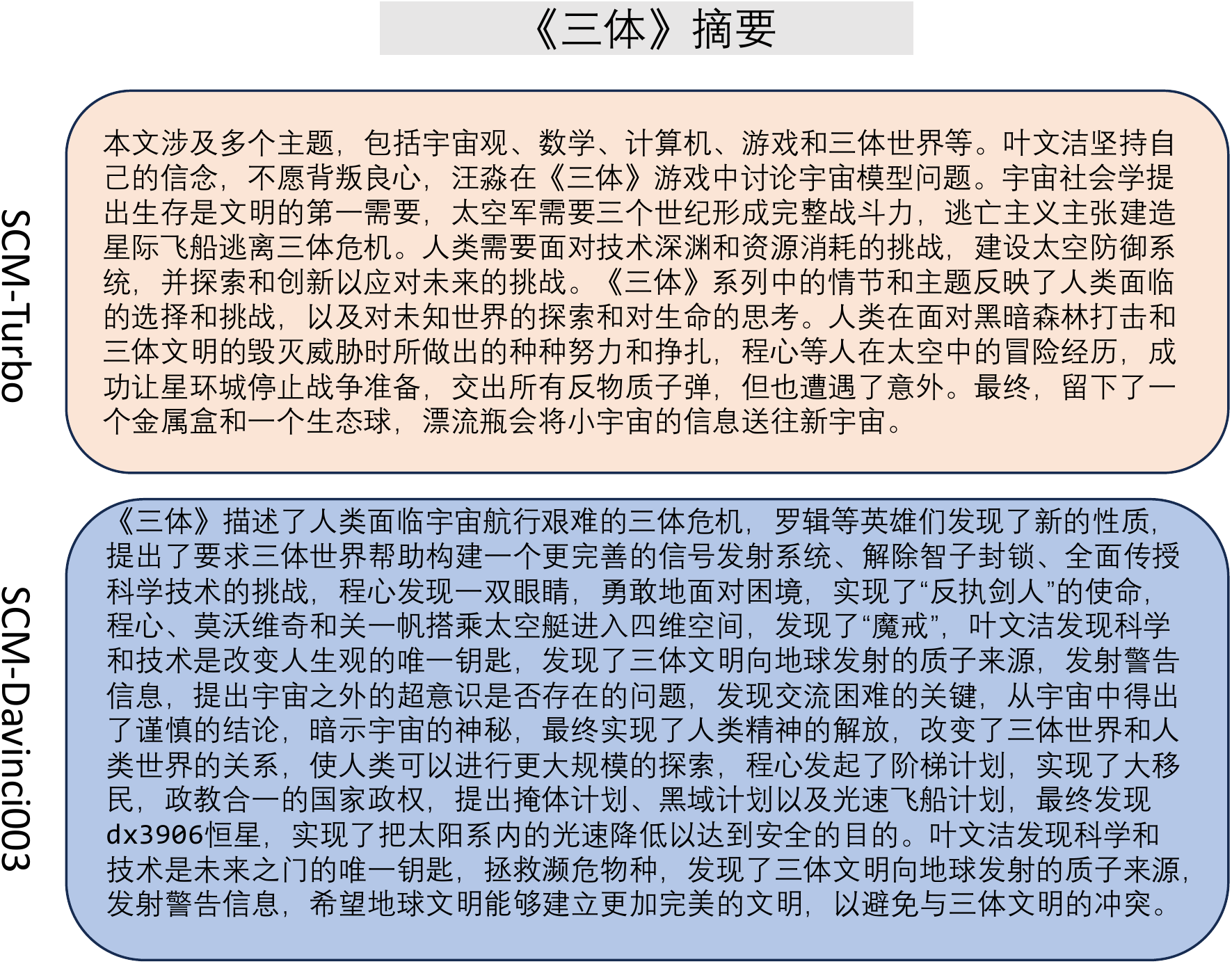}
    \caption{Summarization of the Chinese book \textit{Three Body}.}
\end{figure*}

\begin{figure*}[ht]
    \centering
    \includegraphics[width=0.90\textwidth]{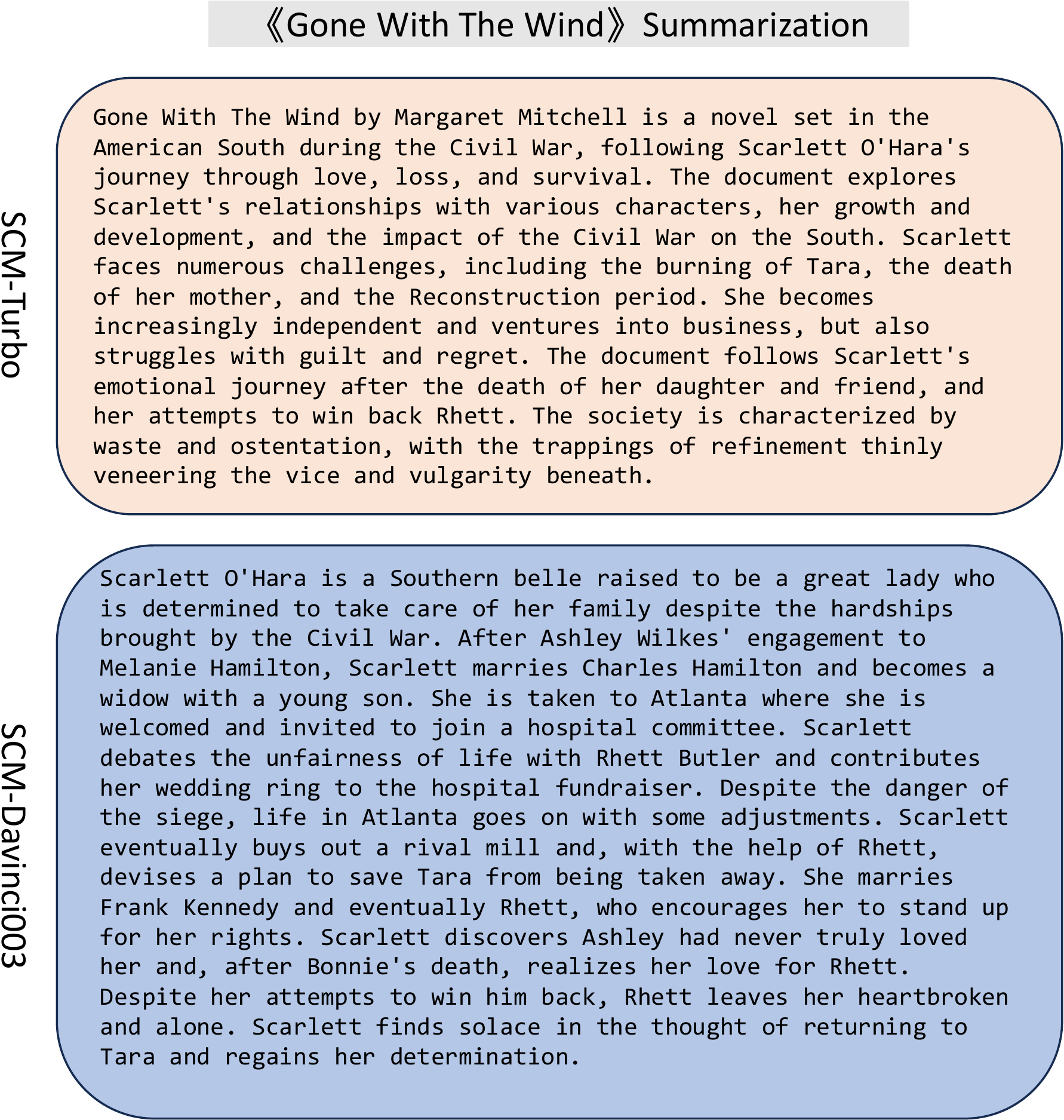}
    \caption{Summarization of the English book \textit{Gone With The Wind}.}
\end{figure*}

\section{Meeting Summarization Cases}
\label{sec:appd-meetingsum-cases}

\begin{figure*}[ht]
    \centering
    \includegraphics[width=0.90\textwidth]{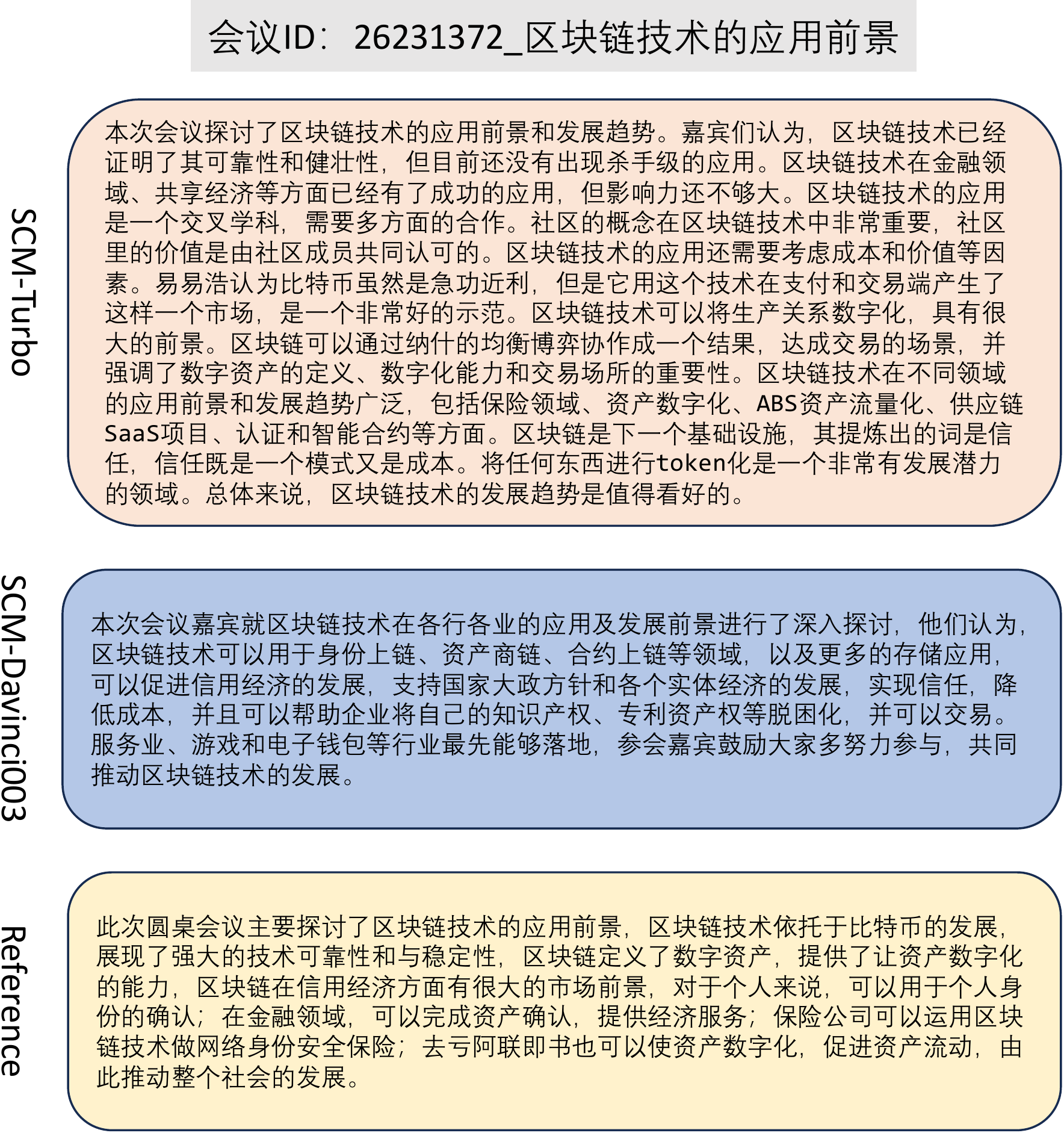}
    \caption{Summarization of the meeting about block chain.}
\end{figure*}

\end{document}